\documentclass[runningheads]{llncs}
\usepackage{url}
\usepackage[T1]{fontenc}     
\usepackage[utf8]{inputenc}  
\usepackage{microtype}       
\usepackage{comment}
\usepackage{times}       
\usepackage{enumitem}        
\usepackage{setspace}        
\usepackage{amsmath}        
\usepackage{amssymb}        
\usepackage{mathtools}       
\usepackage{graphicx}
\usepackage{hyperref}        
\usepackage{caption}         
\usepackage{subcaption}      
\usepackage{booktabs}        
\usepackage{tabulary}        
\usepackage{tabularx}        
\usepackage{multirow}        
\usepackage{longtable}       
\usepackage{array}           
\usepackage{threeparttable}

\usepackage{xcolor}

\definecolor{dscolor}{RGB}{186,29,143}

\usepackage{cite}
\usepackage{inconsolata}
\begin{document}

\title{On Reasoning Behind Next Occupation Recommendation}
\author{Shan Dong\inst{1} \and
Palakorn Achananuparp\inst{1} \and
Hieu Hien Mai\inst{1} \and
Lei Wang\inst{1} \and
Yao Lu\inst{2} \and
Ee-Peng Lim\inst{1}\thanks{Corresponding author.}
}

\authorrunning{S. Dong et al.}

\institute{Singapore Management University \\
\email{\{sdong, palakorna, hhmai, lei.wang.2019, eplim\}@smu.edu.sg}
 \and
Columbia University \\
\email{y12479@columbia.edu}}
\maketitle

\begin{abstract}
In this work, we develop a novel reasoning approach to enhance the performance of large language models (LLMs) in future occupation prediction.  In this approach, a reason generator first derives a ``reason'' for a user using his/her past education and career history. The reason summarizes the user's preference and is used as the input of an occupation predictor to recommend the user's next occupation.  This two-step occupation prediction approach is, however, non-trivial as LLMs are not aligned with career paths or the unobserved reasons behind each occupation decision. We therefore propose to fine-tune LLMs improving their reasoning and occupation prediction performance. We first derive high-quality oracle reasons, as measured by factuality, coherence and utility criteria, using a LLM-as-a-Judge. These oracle reasons are then used to fine-tune small LLMs to perform reason generation and next occupation prediction. Our extensive experiments show that: (a) our approach effectively enhances LLM's accuracy in next occupation prediction making them comparable to fully supervised methods and outperforming unsupervised methods; (b) a single LLM fine-tuned to perform reason generation and occupation prediction outperforms two LLMs fine-tuned to perform the tasks separately; and (c) the next occupation prediction accuracy depends on the quality of generated reasons. Our code is available at \url{https://github.com/Sarasarahhhhh/job_prediction}.
\end{abstract}

\section{Introduction}
\label{sec:intro}

\noindent\textbf{\bf Motivation.}
Understanding and predicting individuals’ career trajectories has significant implications for workforce analytics and career guidance. 
In real-world applications, this capability directly benefits job seekers by providing them with actionable and transparent career transition strategies. Furthermore, at a macro level, analyzing these trajectories aids organizations and policymakers in workforce planning and understanding labor market dynamics.
Since a person's career path can be viewed as a temporal sequence of occupations, fully supervised methods using neural sequential recommendation models (e.g., SASRec \cite{kang2018self} and BERT4Rec \cite{sun2019bert4rec}) can be directly used. However, these models focus on matching the representation of an input history sequence with that of next item, instead of the rationale behind the prediction. In contrast, \textit{reasoning-augmented} large language models (LLMs) that generate step-by-step\cite{wei2022chain} reasons for their predictions, offering a more human-interpretable approach. 
By producing clear rationales for occupational transitions, these models combine predictive power with interpretability, enabling transparent and actionable decision-making.

\noindent\textbf{\bf Objective and Challenges.}
We therefore investigate whether reasoning-augmented language models can generate coherent, plausible, and verifiable reasoning paths that lead to accurate next-occupation predictions. We focus on predicting the next occupation rather than specific jobs, as successfully transitioning into a particular job involves numerous external factors -- such as the candidate pool, hiring processes, and labor market conditions -- over which individuals have little control, making them less useful for personal career decisions. Occupation-level prediction, in contrast, represents a strategic and actionable step: individuals can use the predicted occupation as guidance for skill development or training to facilitate career transitions. Furthermore, focusing on occupations allows us to leverage the structured O*NET-SOC taxonomy, a standardized occupational classification from the U.S. Department of Labor, providing a standardized evaluation framework. 

For reasoning-based approaches, the main challenge is producing rationales that are logically consistent with a user’s career history and contribute substantively to prediction, rather than serving as post-hoc justifications.  Nevertheless, these rationales are often not observed in the data, posing a major challenge to the training and evaluation of reasoning-based methods.  Another challenge is the training of resource efficient small LLMs to generate high quality rationales for accurate reasoning-augmented occupation prediction. 

\noindent\textbf{\bf Next Occupation Prediction Task.}  
Let $\mathcal{E}$ and $\mathcal{J}$ denote the set of all education and job records respectively. 
The job and education history (i.e., user \textit{history}) of a user $u$ can be represented as a sequence of chronologically ordered occupation and education records:
\begin{align}
\mathcal{H}_u = \{h_{u,1}, h_{u,2}, \dots, h_{u,T}\}, \quad h_{u,t} \in \mathcal{E} \cup \mathcal{J},
\end{align}
where $h_{u,t}$ contains attributes such as school name, degree, major, and graduation year if it is an education record, and job title, occupation, industry, time span, and salary if it is a job record. 
The objective of next occupation prediction is to predict the occupation $y_{u,T+1}$ of the next job record based on the user history $\mathcal{H}_u$~\cite{decorte2023career,zheng2023generative,zhang2021attentive}.
Formally, the prediction function is defined as:
%\begin{align}
$\hat{y}_{u,T+1} = \arg\max_{y \in \mathcal{Y}} \; \mathbb{P}(y \mid \mathcal{H}_u)$
%\end{align}
where $\mathcal{Y}$ denotes the set of standardized occupation titles from the O*NET-SOC 2019 taxonomy. 

\noindent\textbf{\bf Contributions.}
Our work makes three main contributions:
(1) We propose a reasoning-augmented framework for next occupation prediction that integrates a reason generator and an occupation predictor, instantiated in both a two-model and a joint-model design. 
(2) We construct high-quality oracle reasons by combining multi-attempt LLM generation with a rationality-based LLM-as-a-Judge filter that evaluates factuality, coherence, and utility.
(3) We develop effective fine-tuning strategies, including SFT and DPO, and show that reasoning-augmented models improve prediction accuracy, the joint model performs best, and prediction quality depends strongly on reason quality.

\section{Related Work}

We may categorize the related work into sequential recommendation and large language model-based approaches.  In the former approach, a deep sequence model such as RNN, GRU, or transformer is trained to predict the target item given an input sequence of earlier items. 
Paparrizos et al.  formulated a supervised machine learning problem using features extracted from job transitions, employees and employers~\cite{paparrizos2011machine}. Instead of predicting the next job, Yamashita et al. proposed to predict the future job sequence~\cite{yamashita2022looking} using a transformer model trained with the job and company representations derived from job and company transition graphs, logical rules about job sequences, and job title embedding.  The above work, however, does not provide textual reasons to explain why a user should consider the predicted future job or career. This weakens the trustworthiness of recommendation and may reduce the adoption of recommendations.

LLMs in recent years have demonstrated its strong ability to solve NLP and mathematical reasoning tasks~\cite{achiam2023gpt, gemini2023, openai2025omni}. Researchers hence started to explore using LLMs to perform prediction/recommendation. Zhang et al. proposed to train LLM to perform recommendation as an instruction following task \cite{zhang2025recommendation}. Their method, InstructRec, fine-tunes a LLM with recommendation instructions constructed from interaction histories using different instruction templates.  
This method, however, has not been applied to job recommendation.  Moreover, it lacks the reasoning aspect very useful in career guidance. 

In job recommendation, Wu et al. proposed recommendation task-specific prompts to predict if a user likes or dislikes a candidate job in a point-wise instruction, and to predict if  a user prefers a candidate job over another candidate in a pair-wise instruction~\cite{wu2024exploring}. 
The approach's efficiency nevertheless suffers when there is a large number of candidate jobs.  
In \cite{du2024enhancing}, an LLM-based job recommendation method GANs Interactive Recommendation (LGIR) was proposed to improve the poor quality career trajectory information with the help of Generative Adversarial Networks so as to achieve better recommendation results.  
The above LLM-based methods, however, do not consider reasoning as an approach to enhance prediction accuracy as well as to explain why an occupation should be recommended to the user based on his/her education and job history. 

\section{Preliminary}

\noindent\textbf{\bf Dataset and Data Preprocessing.}
% \label{subset:data_source}
We conduct our research on a proprietary resume dataset from Lightcast\footnote{https://lightcast.io/products/data/overview}. 
To ensure data reliability, we apply minimal yet essential preprocessing.
First, job records with missing start dates or unclassified occupation names are removed.
Second, only users with 5–15 valid jobs are retained for adequate trajectory length.
Education records are always preserved, even when certain date fields are missing. 
We initially select the education and job records of over 11,000 US white-collar workers. Each job record is enriched with standardized occupational titles and 8-digit codes aligned with the O*NET-SOC 2019 taxonomy\footnote{https://www.onetcenter.org/taxonomy/2019/list.html}. We first hold out 1,000 users as the test set. For the remaining data, we perform our data filtering procedure and obtain 3,646 high-quality users for training, resulting in a final dataset of 4,646 users.

\begin{figure*}[!t]
    \centering
    \includegraphics[width=\textwidth]{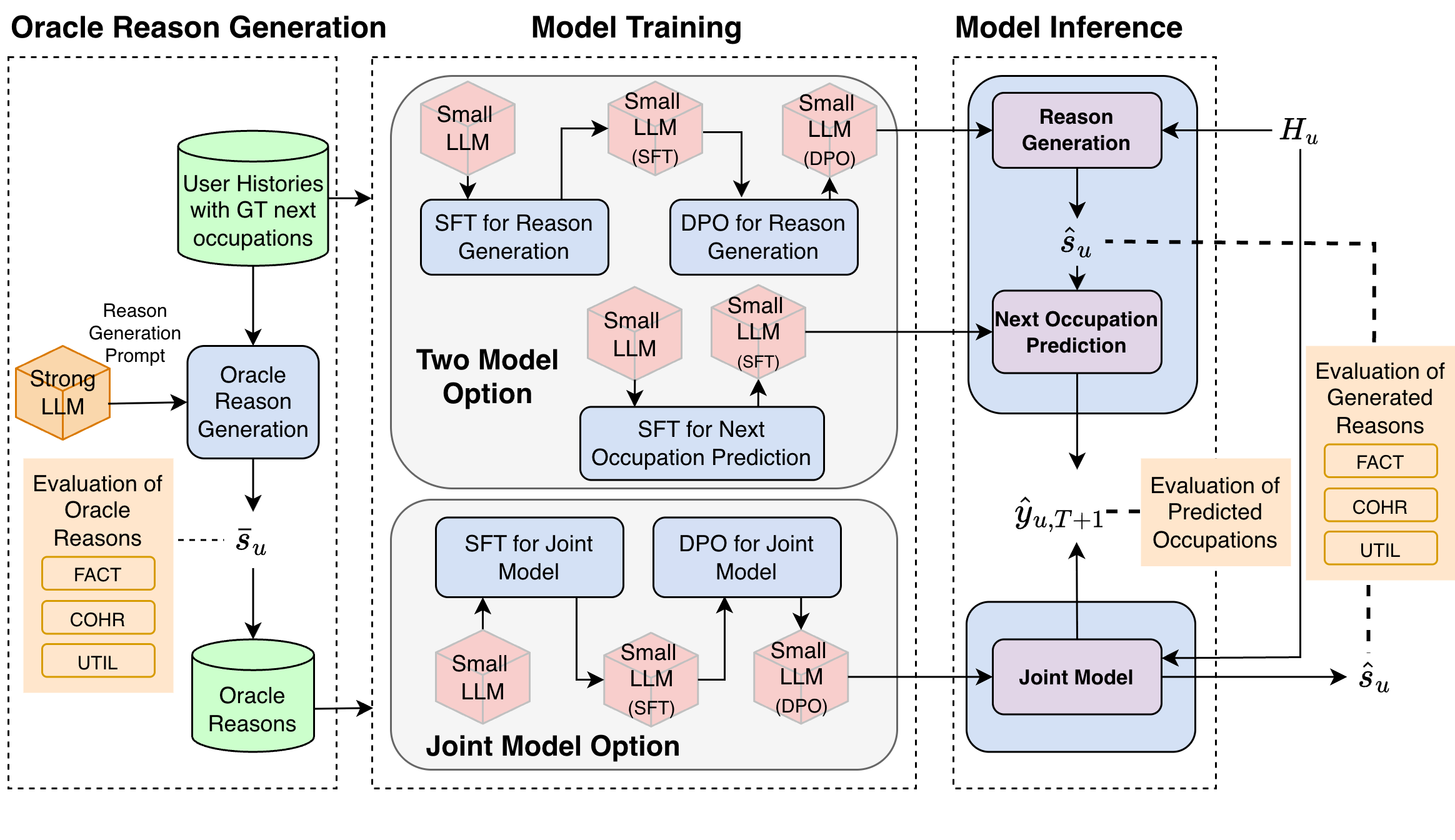} 
    \vspace{-10mm}
    \caption{Reasoning-Augmented Occupation Prediction Framework.}
    \label{fig:overall_framework}
    \vspace{-6mm}
\end{figure*}

\noindent\textbf{\bf Proposed Framework.}
Figure~\ref{fig:overall_framework} illustrates our proposed framework of reasoning-augmented next occupation prediction. It consists of three major phases: 
\textbf{oracle reason generation}, \textbf{model training}, and \textbf{model inference}. 
In the oracle reason generation phase, we build a dataset that connects each user's education and career history with the target next occupation using a high-quality explanation text, called the \textit{oracle reason}. 
We generate oracle reasons because ground truth reasons of users selecting their next occupations are rarely captured in real world data. In our framework, we prompt a strong LLM to generate reasons and introduce a few novel filtering strategies and rationality-based evaluation methods to retain the high quality ones for model training.  
In the model training stage, we train small LLM(s) to infer reasons and predict the next occupation using our constructed user history and next occupation data.
The training process include two stages, namely:
(1) \textbf{Supervised Fine-Tuning (SFT)} that teaches the model to both generate reasons and predict next occupations;
(2) \textbf{Direct Preference Optimization (DPO)}~\cite{rafailov2023direct, liu2025survey} that refines reason generation quality by aligning the model output with preferred reasons.
In the model inference phase, the fine-tuned model is used to predict the next occupation for unseen users.
The model takes an unseen user’s education and job history as input, generates a reasoning explanation, and subsequently predicts the most likely next occupation.

\section{Oracle Reason}
\label{sec:oracle_reason}
\noindent\textbf{\bf Oracle Reason Generation.}
A user may have many possible motivations when seeking their next occupation. The most likely reason is one that reflects the user’s preference and how this preference guides the choice of next occupation. We thus construct an oracle reason by leveraging both the user’s career history and the ground-truth next occupation.
Specifically, we provide a strong LLM (GPT-4.1) with the user’s career history $\mathcal{H}_u$, and prompt it to jointly generate both the reason and the next occupation:
\vspace{-2mm}
\begin{align}
\langle \bar{\mathbf{s}}_{u}, \tilde{y}_{u,T+1} \rangle = \text{LLM}(\mathcal{H}_u),
\end{align}
where $\bar{\mathbf{s}}_{u}$ denotes the generated oracle reason and $\tilde{y}_{u,T+1}$ is the predicted next occupation.
The model may not always predict the correct next occupation. To obtain high-quality reasons while still keeping a sufficient number of samples, we adopt strategy below:

\begin{itemize}[leftmargin=*]
\item \textbf{Multiple-Attempt Generation:} For each user, the LLM generates multiple independent $\langle \text{reason}, \text{prediction} \rangle$ pairs.
\item \textbf{Correct-Prediction Filtering:} Generated reasons with correct predictions (\(\tilde{y}_{u,T+1} = y_{u,T+1}\)) are retained as valid reasons.
\item \textbf{User-Level Retention and Sampling:} 
A user is included in the training set only if at least one valid reason exists.
For each retained user, we randomly sample one valid reason to create a single training instance.
\end{itemize}
We denote this training dataset by $
\mathcal{D}_{\text{trg}} 
  = \Bigl\{ \bigl(\mathcal{H}_u,\bar{\mathbf{s}}_{u}, y_{u,T+1}\bigr) \;\Big|\; \tilde{y}_{u,T+1} = y_{u,T+1} \Bigr\}.
$

Inspired by~\cite{tsai2024leveraging, lee2025evaluating}, we evaluate the quality of generated reasons based on their rationality along three dimensions: 
\textit{Factuality}, \textit{Coherence}, and \textit{Utility}. 
An \textbf{LLM-as-a-Judge} paradigm~\cite{zheng2023judging} (GPT-4.1) is adopted, where a strong LLM provides dimension-specific scores based on the input user history, the reason text, and the ground-truth next occupation.

\noindent\textbf{Factuality ({\sc Fact}).}
Factuality measures whether the reasoning content aligns with the given user history.
It requires that all stated education or job information in the reason can be grounded in the input history, without fabrication or omission of essential facts.

\noindent\textbf{Coherence ({\sc Cohr}).}
Coherence assesses the logical consistency and structure of the reasoning narrative.
It checks that each step follows a clear temporal or causal order.

\noindent\textbf{Utility ({\sc Util}).}
Utility evaluates whether the reasoning effectively supports the next-occupation decision.
It considers whether the reasoning provides relevant evidence and contributes directly to justifying the  ground-truth next occupation.

\noindent\textbf{Scoring with LLM-as-a-Judge.}
For each reason $\hat{\mathbf{s}}$, we use a strong LLM as a judge and prompt it to output a rationality score $g_{d}(\hat{\mathbf{s}}) \in [1,5]$ for each dimension 
$d \in \{\textsc{Fact}, \textsc{Cohr}, \textsc{Util}\}$, along with brief justifications.
A reason is retained only if it satisfies the per-dimension threshold:
$g_{d}(\hat{\mathbf{s}}) \ge 4.0 \quad \text{for all } d$.
Only $3,646$ samples passing this criterion are retained as valid oracle reasoning triplets for training.

\noindent\textbf{\bf{Evaluation of LLM-as-a-Judge via Perturbation.}}
We evaluate the robustness of the LLM-as-a-Judge using 100 examples, applying controlled perturbations to reasoning along three dimensions: factuality, coherence, and utility, and then measuring the corresponding score changes. We introduce minor and major perturbations by modifying 1 vs.\ 2--3 factual items, shuffling 20\% vs.\ 50\% of sentences to disrupt coherence, and replacing the ground-truth next occupation with a related vs.\ unrelated one to alter utility.

\begin{table*}[!t]
\centering

\caption{
\textbf{LLM-as-a-Judge evaluation results.}
Average scores (1–5) across factuality, coherence, and utility under oracle, minor, and major perturbations.
}

\footnotesize
\begin{tabular}{l|ccc}
\toprule
\textbf{Reason Type} & \textbf{Factuality} & \textbf{Coherence} & \textbf{Utility} \\
\midrule
Oracle Reason & 4.85 & 4.90 & 4.93 \\
Minor Perturbation & 3.30 & 3.96 & 3.10 \\
Major Perturbation & 2.80 & 2.10 & 1.37 \\
\bottomrule
\end{tabular}
\label{tab:llm-as-judge-reason-eval}
\vspace{-7mm}
\end{table*}

As shown in Table~\ref{tab:llm-as-judge-reason-eval}, the judge is highly sensitive to these perturbations. Oracle reasoning receives the highest scores. Minor perturbations cause moderate declines, while major perturbations lead to substantial drops across all dimensions, indicating that the evaluator reliably identifies degraded or misleading reasoning.

\section{Model Training and Inference}

\paragraph{\bf Supervised Fine-Tuning (SFT).}
Using the high-quality triplets of (user history, reason, ground truth next occupation) in $\mathcal{D}_{\text{trg}}$, we use SFT to fine-tune models to generate reasons and to predict next occupations.  We consider two model options below.

\noindent\textbf{Two-Model Option.}  In this option, we finetune two LLMs, \textit{Reason Generator} $\text{LLM}_{\theta_{rg}}$ and \textit{Occupation Predictor} $\text{LLM}_{\theta_{pred}}$ for reason generation and occupation prediction respectively.
Let $\theta_{rg}$ be the parameters of the reason generator. We fine-tune  $\text{LLM}_{\theta_{rg}}$ to take a user history $\mathcal{H}_u$ as input and  generate the reason $\bar{\mathbf{s}}_u$. The training loss is the negative log-likelihood of the reasoning tokens:
\vspace{-2mm}
\begin{align}
\mathcal{L}_{rg} 
= - \sum_{u} \log \mathbb{P}_{\theta_{rg}}\bigl(\bar{\mathbf{s}}_{u} \mid \mathcal{H}_u \bigr),
\end{align}

Let $\theta_{pred}$ denote the parameters of the occupation predictor. We next fine-tune $\text{LLM}_{\theta_{pred}}$ to predict the ground-truth next occupation $y_{u,T+1}$ conditioned on both the user history $\mathcal{H}_u$ and  $\bar{\mathbf{s}}_u$. The training loss is the cross-entropy over occupation labels:
\vspace{-2mm}
\begin{align}
\mathcal{L}_{pred}=  - \sum_{u} \log \mathbb{P}_{\theta_{pred}}\bigl(y_{u,T+1} \mid \mathcal{H}_u, \bar{\mathbf{s}}_u \bigr),
\end{align}

\noindent\textbf{Joint Model Option.}  Here, we fine-tune only one model $\text{LLM}_{\theta_{comb}}$ to generate the reason and next occupation simultaneously by optimizing the joint loss function $\mathcal{L}_{rg}+\mathcal{L}_{pred}$. 

\paragraph{\bf Direct Preference Optimization (DPO)}
We further train the reasoning model using DPO~\cite{rafailov2023direct}, which further aligns the model’s reasoning outputs with generated high-quality reasons.

\noindent\textbf{Data Preparation.}
For data ($\mathcal{D}_{\text{trg}}$), we can naturally construct \textit{preference pairs} consisting of a preferred (positive) and a non-preferred (negative) reasoning example:
\begin{itemize}[leftmargin=*]
    \item \textbf{Positive sample:} $(\mathcal{H}_u, \bar{\mathbf{s}}_u^+, y_{u,T+1})$, where the generated reasoning $\bar{\mathbf{s}}_u^+$ correctly leads to the ground-truth next occupation $y_{u,T+1}$.
    \item \textbf{Negative sample:} $(\mathcal{H}_u, \bar{\mathbf{s}}_u^-, \tilde{y}_{u,T+1})$, where the reasoning $\bar{\mathbf{s}}_u^-$ leads to an incorrect prediction $\tilde{y}_{u,T+1} \neq y_{u,T+1}$.
\end{itemize}

Given such reasoning pairs $(\bar{\mathbf{s}}_u^+, \bar{\mathbf{s}}_u^-)$ conditioned on the same user history $\mathcal{H}_u$, DPO directly optimizes the model $\pi_\theta$ to increase the likelihood of preferred (positive) reasoning while decreasing that of the non-preferred (negative) one:

\begin{align}
\mathcal{L}_{\text{DPO}}
&= - \log \sigma \Bigl(
  \beta \Bigl[
    \log \pi_\theta(\bar{\mathbf{s}}_u^+ \mid \mathcal{H}_u)
    - \log \pi_\theta(\bar{\mathbf{s}}_u^- \mid \mathcal{H}_u) \nonumber \\
& \quad
    - \log \pi_{\text{ref}}(\bar{\mathbf{s}}_u^+ \mid \mathcal{H}_u)
    + \log \pi_{\text{ref}}(\bar{\mathbf{s}}_u^- \mid \mathcal{H}_u)
  \Bigr]
\Bigr),
\end{align}
where $\pi_{\text{ref}}$ is the reference model (e.g., the SFT model), $\sigma(\cdot)$ is the sigmoid function, and $\beta$ controls the sharpness of preference alignment. 
This objective encourages $\pi_\theta$ to assign higher probability to human-preferred reasoning $\bar{\mathbf{s}}_u^+$ than to non-preferred reasoning $\bar{\mathbf{s}}_u^-$, effectively refining reasoning quality without explicitly computing reward gradients.

For the joint model $\text{LLM}_{\theta_{comb}}$, DPO is directly applied to the joint generation sequence $(\bar{\mathbf{s}}_u, y_{u,T+1})$.  
In the Two-Model setting, DPO is applied to the reasoning generator $\text{LLM}_{\theta_{rg}}$ while keeping the occupation predictor $\text{LLM}_{\theta_{pred}}$ frozen.  
Each DPO stage continues training from its corresponding SFT checkpoint to ensure stability and efficiency.

\paragraph{\bf Model Inference.}
At inference time, given an unseen user history $\mathcal{H}_u$, the fine-tuned recommendation model(s) is applied to generate reason and to predict the next occupation.
\begin{itemize}
  \item \noindent\textbf{Two-Model Option.} In the first step, the reason generator $\text{LLM}_{\theta_{rg}}$ produces a reasoning text $\hat{\mathbf{s}}_u$ conditioned only on the user history.
    The occupation predictor $\text{LLM}_{\theta_{pred}}$ then takes both $\mathcal{H}_u$ and the generated reasoning $\hat{\mathbf{s}}_u$ as input, and is trained to output the predicted next occupation.
\vspace{-2mm}
\begin{align}
\hat{\mathbf{s}}_u&=\text{LLM}_{\theta_{rg}}(\mathcal{H}_u)\\
\hat{y}_{u,T+1}&=\text{LLM}_{\theta_{pred}}(\mathcal{H}_u,\hat{\mathbf{s}}_u)
\end{align}

\item \noindent\textbf{Joint-Model Option.}  We obtain the generated reason and predicted next occupation from $\text{LLM}_{\theta_{comb}}$ in one step. That is:
\vspace{-2mm}
\begin{align}
(\hat{\mathbf{s}}_u,\hat{y}_{u,T+1})=\text{LLM}_{\theta_{comb}}(\mathcal{H}_u)
\end{align}
\end{itemize}

\section{Experiments}
\label{sec:expt}

\paragraph{\bf Data Preparation.} 
After preprocessing, we obtain over 11,000 user records and hold out one 1,000-user test set. The remaining users form the training pool. Our multi-stage pipeline (Section~\ref{sec:oracle_reason}) produces 3,646 high-quality oracle examples $\mathcal{D}_{\text{trg}}$, which we use as the SFT dataset $\mathcal{D}_{\text{SFT}}$. For DPO, we exclude users whose multiple LLM predictions are all correct, since constructing preference pairs requires at least one incorrect prediction. This yields 2,651 users with valid positive--negative reasoning pairs, forming $\mathcal{D}_{\text{DPO}}$.

\noindent\textbf{Model Training.}
We adopt \textsf{Qwen3-8B}~\cite{yang2025qwen3} as the backbone for all experiments. 
Two training paradigms are considered: a \textit{separate training} setup that independently trains a reason generator and an occupation predictor, and a \textit{joint training} setup that produces both the reason and the next occupation prediction together. 
We train models across two stages: \textbf{Supervised Fine-Tuning (SFT)} and \textbf{Direct Preference Optimization (DPO)}. 
All fine-tuning is performed with \textbf{full-parameter training}.

\textbf{SFT stage.}
For the two model option (\textsf{Qwen3-8B-Two-Model-SFT}), we obtain two models per dataset:   
(a) a reason generator and (b) an occupation predictor, denoted as  
\textsf{Qwen3-8B-Two-Model-SFT-R} and \textsf{Qwen3-8B-Two-Model-SFT-P}.  
For the joint paradigm, we train \textsf{Qwen3-8B-Joint-SFT}.  
We use LLaMa-Factory~\cite{zheng2024llamafactory} to fine-tune with a maximum sequence length of 2{,}048 tokens, batch size $8$, learning rate $2\times10^{-5}$, cosine learning rate schedule, $8$ epochs, \texttt{bf16} precision, and DeepSpeed ZeRO-3.

\textbf{DPO stage.}
DPO further aligns the generated explanations with the oracle reasons.  
For the two model option (\textsf{Qwen3-8B-Two-Model-SFT-DPO}), only the reason generator is updated, producing  
\textsf{Qwen3-8B-Two-Model-SFT-DPO-R},  
while the occupation predictor remains frozen.  
For the joint paradigm, we obtain \textsf{Qwen3-8B-Joint-SFT-DPO}.  
DPO uses $\beta{=}0.1$ (sigmoid DPO loss), maximum sequence length $2{,}048$, batch size $2$ with gradient accumulation of $4$, learning rate $5\times10^{-6}$, and $5$ training epochs.

\paragraph{\bf Baselines}
We compare our proposed reasoning-augmented models with two types of baselines: 
\begin{itemize}
    \item \textbf{Transformer-based Recommenders.}: We adopt two strong baseline Transformer-based sequential recommendation models -- \textbf{SASRec} \cite{kang2018self} and \textbf{BERT4Rec} \cite{sun2019bert4rec}.

For both training and test users, we use the observed history occupations to predict the next occupation. 
  \item \textbf{Direct SFT (No Reason).} We include two supervised fine-tuned models, namely LLaMA-3.1-8B-Direct-SFT and Qwen3-8B-Direct-SFT. These models are trained to directly predict the next occupation based on the user history.
  \item \textbf{Zero-shot CoT methods.} They include zero-shot CoT methods \cite{kojima2022large, wang2023zero} on \textbf{LLaMA-3.1-8B}~\cite{grattafiori2024llama}, \textbf{Qwen3-8B}~\cite{yang2025qwen3}, and \textbf{GPT-4.1}~\cite{achiam2023gpt}. These baseline methods are unsupervised and they only use a user's education and job history as input, generate a reason and predict the next occupation in a single prompt.
\end{itemize}

We evaluate the quality of predicted next occupations using two complementary metrics: \textit{Exact Match Accuracy} and \textit{Related Occupation Match Accuracy}.

\noindent\textbf{Exact Match Accuracy.}
Let $\hat{y}_{u,T+1}$ denote the predicted occupation title for user $u$ and $y_{u,T+1}$ denote the ground-truth occupation title (standardized to O*NET-SOC 2019). Let $\mathbf{I(\cdot)}$ be an indication function which returns 1 if the input is TRUE, and 0 otherwise.  
Exact Match Accuracy is then defined as the proportion of users in the test set $\mathcal{U}_{test}$ having their next occupations predicted correctly:
\begin{align}
\mathrm{Acc}_{\text{EM}} = \frac{1}{|\mathcal{U}_{test}|} \sum_{u \in \mathcal{U}_{test}} \mathbf{I}(\hat{y}_{u,T+1} = y_{u,T+1}).
\end{align}

\noindent\textbf{Related Occupation Match Accuracy.}
To account for multiple plausible next occupations, we evaluate whether the predicted occupation matches any occupation related to the ground truth according to the O*NET database. For each ground-truth occupation $y_{u,T+1}$, we define a ranked list of related occupations:

\begin{align}
\mathcal{R}(y_{u,T+1}) = \bigl[ r_1, r_2, \dots, r_K \bigr],
\end{align}
where $r_1 = y_{u,T+1}$ (the ground truth itself), and subsequent $r_k$ are sorted related occupations provided by O*NET Resource Center.  
Given a prediction $\hat{y}_{u,T+1}$, we define the \textit{Related Occupation Match Accuracy} as the average of reciprocal rank of $\hat{y}_{u,T+1}$ in $\mathcal{R}(y_{u,T+1})$:
\begin{align}
\mathrm{Acc}_{\text{RM}} 
= \frac{1}{|\mathcal{U}_{test}|} \sum_{u \in \mathcal{U}_{test}} \frac{1}{k} \cdot \mathbf{I}(\hat{y}_{u,T+1} = r_k).
\end{align}

This metric rewards predictions that are relevant to the ground-truth occupation. A perfect exact match yields $\frac{1}{k}=1$, while related but not identical occupations yield fractional scores decreasing with rank position.

\begin{table*}[!t]
\centering
\caption{Occupation Prediction Accuracy on 1,000 test samples. 
}
\footnotesize
\resizebox{0.68\linewidth}{!}{
\begin{tabular}{l l | c c }
\hline
\textbf{Model Category} & \textbf{Model} 
 &  \textbf{$\mathbf{Acc}_{\mathbf{EM}}$} & \textbf{$\mathbf{Acc}_{\mathbf{RM}}$} \\
\hline

% ---------------- Traditional Rec Baselines ----------------
% \multirow{2}{*}{\textbf{Traditional Recommenders}} 

\textbf{Transformer} & SASRec   & 25.30\% &  29.83\% \\
\textbf{Recommenders} & BERT4Rec & 23.60\% & 28.82\% \\
\hline

% ---------------- Non-training LLMs ----------------
\multirow{3}{*}{\textbf{Zero-shot CoT}} 
 & LLaMA-3.1-8B  & 6.70\%  & 8.13\%  \\
 & Qwen3-8B      & 24.60\% & 27.87\%  \\
 & GPT-4.1       & 23.90\% & 28.86\%  \\
\hline

\multirow{2}{*}{\textbf{No Reasoning}}
 & LLaMA-3.1-8B-Direct-SFT  & 25.50\%  & 30.47\% \\
 & Qwen3-8B-Direct-SFT       & 26.70\% & 31.92\%  \\
\hline

% ---------------- Training ----------------
\multirow{8}{*}{\textbf{Proposed Models}}
  & LLaMa-3.1-8B-Two-Model-SFT         & 26.10\% & 31.12\% \\
 & LLaMa-3.1-8B-Two-Model-SFT-DPO         & 26.70\% & 31.62\% \\
 \cline{2-4}
  & Qwen3-8B-Two-Model-SFT           & 27.40\% & 32.24\%  \\
  & Qwen3-8B-Two-Model-SFT-DPO$^{\dagger}$     & 31.00\% & 35.35\%  \\
   \cline{2-4}
 & LLaMA-3.1-8B-Joint-SFT       & 26.90\% & 31.88\%  \\
 & LLaMA-3.1-8B-Joint-SFT-DPO   & 27.60\% & 32.85\%  \\
  \cline{2-4}
 & Qwen3-8B-Joint-SFT           & 28.20\% & 32.83\%  \\
 & Qwen3-8B-Joint-SFT-DPO$^{\dagger}$       & 31.40\% & 36.46\%  \\

\hline

\end{tabular}
}

\begin{tablenotes}
\footnotesize
\item $^{\dagger}$ means that these two DPO models significantly outperform all other models according to McNemar’s test \cite{mcnemar1947note} with Bonferroni correction.
\end{tablenotes}
\vspace{-5mm}
\label{tab:main_results}
\end{table*}

\section{Results and Discussion}
\noindent\textbf{Overall Performance.}
Table~\ref{tab:main_results} presents the overall next-occupation prediction accuracy across traditional recommenders, zero-shot CoT, and our reasoning-augmented models. Transformer-based sequential recommenders (SASRec and BERT4Rec) are strong supervised baselines as they are trained directly on career sequences. The results show that zero-shot CoT methods, with LLMs' powerful pretrained knowledge, could not outperform these supervised methods. %Not surprisingly, zero-shot CoT with GPT-4.1 delivers the best results. 
These CoT methods lack exposure to task-relevant training signals and thus struggle to generate reasons that accurately reflect users’ career histories and align with the prediction task.

Our proposed models with fine-tuning delivers substantially improved accuracy. Specifically, the joint model option with SFT only outperforms all zero-shot CoT methods,
indicating that training reasoning and prediction jointly improves both interpretability and performance.
DPO brings a further performance boost by aligning the generated reasons with the oracle reasons.
Our best model, Qwen3-8B-Joint-SFT-DPO, achieves the highest accuracy in both 
\textsc{Acc\textsubscript{EM}} and \textsc{Acc\textsubscript{RM}}, demonstrating that 
high-quality reason is crucial for accurate occupation prediction.
Notably, the Direct-SFT baselines outperform most zero-shot CoT methods, indicating that supervised fine-tuning alone provides strong gains. However, our reason-augmented SFT models consistently achieve further improvements, demonstrating the additional benefit of explicit reason generation beyond direct prediction.
Interestingly, GPT-4.1, despite its larger scale, achieves a slightly lower exact-match accuracy than Qwen3-8B. 

\noindent\textbf{Two-Model vs Joint-Model options.}
Table~\ref{tab:main_results} compares the Two-Model and Joint-Model training strategies.
Joint-Model training consistently outperforms the Two-Model approach under both SFT and SFT-DPO.
It benefits from generating the reason and the predicted occupation within a single sequence, allowing the prediction to be directly conditioned on the model’s own reasoning during training. This reduces the inconsistency between output reason and occupation.
In contrast, the Two-Model option separates the reason generator from the occupation predictor.
During training, the predictor learns from oracle reasons.
During inference, the predictor instead relies on generated reasons, which are often less accurate, less detailed, or stylistically different from the oracle.
This mismatch between the training-time and inference-time inputs leads a degradation of prediction quality.

\noindent\textbf{Reasoning Quality and Alignment.}
\begin{table*}[!t]
\centering
\caption{Reasoning quality (by the Rationality-based Evaluation) and text similarity against oracle reason.}
\footnotesize
\resizebox{1.0\linewidth}{!}{
\begin{tabular}{l|cccc|cccc}
\hline
\textbf{Model} & \textbf{Fact.} & \textbf{Cohr.} & \textbf{Util.} & \textbf{Overall} & \textbf{BLEU} & \textbf{ROUGE-1} & \textbf{ROUGE-2} & \textbf{ROUGE-L}  \\
\hline
Oracle Reason   & 4.86 & 4.93 & 4.98 & 4.92 & --     & --     & --     & --      \\
\hline

Zero-shot CoT (LLaMA-3.1-8B) & 4.28 & 4.31 & 4.30 & 4.33 & 0.0916 & 0.5567 & 0.2429 & 0.3137  \\
Zero-shot CoT (Qwen3-8B)  & 4.75 & 3.83 & 4.18 & 4.25 & 0.0463 & 0.3380 & 0.1411 & 0.1912  \\
\hline
Proposed Model (LLaMA-3.1-8B-Joint-SFT)             & 4.68 & 4.61 & 4.35 & 4.55 & 0.0983 & 0.5419 & 0.2187 & 0.2948  \\
Proposed Model (LLaMA-3.1-8B-Joint-SFT-DPO)  & 4.73 & 4.87 & 4.44 & 4.68 & 0.1722 & 0.6494 & 0.3218 & 0.3905 \\
Proposed Model (Qwen3-8B-Joint-SFT)               & 4.75 & 4.93 & 4.57 & 4.75 & 0.1260 & 0.6255 & 0.2919 & 0.3647  \\
Proposed Model (Qwen3-8B-Joint-SFT-DPO)              & 4.76 & 4.94 & 4.70 & 4.80 & 0.1855 & 0.6697 & 0.3363 & 0.4129  \\

\hline
\end{tabular}
}
\label{tab:reason_quality}
\vspace{-6mm}
\end{table*}
Table~\ref{tab:reason_quality} shows the evaluation scores of generated reasons (factuality, coherence, utility) and text similarity (BLEU, ROUGE) between model generated reasons and oracle reasons, based on 100 random test examples.

Our results show that models with DPO generate reasons of higher quality and more similar to oracle reasons compared with models with SFT only and zero-shot CoT methods. This suggests that preference-based alignment reduces both hallucinations (factuality), and enhances structural consistencies (coherence) and relevance to the predicted occupation (utility). 
Qwen3 consistently outperforms LLaMA-3.1 under joint training, suggesting that Qwen3 offers better controllability in generating reasons and achieves closer alignment with oracle reasons.
These above observations directly support our findings in Table~\ref{tab:main_results}: models with more factual, coherent, and utility-aligned reasoning produce more accurate predictions. 
Note that although DPO significantly improves reasoning quality, the gain in next-occupation accuracy over SFT (Table~\ref{tab:main_results}) is relatively small. As shown in Table~\ref{tab:reason_quality}, SFT-generated reasons can also achieve good rationality scores (${>}4.5$ across most dimensions).

\begin{table*}[!t]
\centering
\caption{
Comparison of reason generator and occupation predictor combinations.
}
\label{tab:reason_pred_combo_reversed}
\footnotesize
\resizebox{0.74\linewidth}{!}{
\begin{tabular}{l l c c}
\hline
\textbf{Occupation Predictor (P)} & \textbf{Reason Generator (R)} & \textbf{$\mathbf{Acc}_{\mathbf{EM}}$} & \textbf{$\mathbf{Acc}_{\mathbf{RM}}$} \\
\hline
\multirow{4}{*}{ Qwen3-8B-P } 
    & Oracle Reason          & 83.33\% & 84.09\% \\
    & Qwen3-8B-R              & 61.00\% & 63.82\% \\
    & Qwen3-8B-Two-Model-SFT-R        & 57.67\% & 62.38\% \\
    & Qwen3-8B-Two-Model-SFT-DPO-R    & 63.00\% & 66.62\% \\

\hline
\multirow{4}{*}{ Qwen3-8B-Two-Model-SFT-P } 
    & Oracle  Reason          & 94.67\% & 95.12\% \\
    & Qwen3-8B-R            & 61.00\% & 65.91\% \\
    & Qwen3-8B-Two-Model-SFT-R        & 64.67\% & 69.60\% \\
    & Qwen3-8B-Two-Model-SFT-DPO-R   & 66.33\% & 70.26\% \\
\hline
\end{tabular}
}
\label{tab:combination}
\vspace{-3mm}
\end{table*}

\noindent\textbf{Impact of Reason Quality on Prediction.}
We evaluated 300 test examples with oracle reasons as not all test examples come with oracle reasons. We aim to investigate the impact of oracle and generated reasons, on occupation prediction accuracy under the two-model option of our proposed method using Qwen3-8B (as it is better than LLaMA-3.1-8B in prediction task). 
Table~\ref{tab:combination} compares different combinations of reason generators and occupation predictors and shows that reason quality has a direct impact on prediction performance. Specifically, predictors using oracle reasons yield the highest accuracies. 
Keeping the predictor unchanged, using higher-quality reasons, especially generated by the SFT-DPO reasoner, generally produces better predictions. The predictors trained on oracle reasons have noticeably reduced accuracy when coupled with weaker reasoners. 
Interestingly, the unfinetuned predictor Qwen3-8B-P coupled with Qwen3-8B-R slightly outperforms the same predictor coupled with SFT reasoner. A plausible explanation is that Qwen3-8B-P has never been trained on oracle-style explanations and therefore aligns better with the more free-form, pretrained reasoning style of Qwen3-8B-R. In contrast, the finetuned predictor Qwen3-8B-Two-Model-SFT-P benefits more from Qwen3-8B-Two-Model-SFT-R, whose outputs are closer to the oracle reason distribution used during predictor training.
Overall, the results demonstrate that better reasons lead to better next-occupation predictions.

\noindent\textbf{Scaling Effects.}

We experiment with Qwen3-8B-Joint model and vary the size of the training set from 100 to 1,000 and then to the full dataset. 
EM accuracy improves consistently as more training data is used, for both SFT ($24.2\% \to 26.1\% \to 28.2\%$) and SFT-DPO models ($27.1\% \to 27.8\% \to 31.4\%$). We also compare backbone sizes (1.7B $\to$ 4B $\to$ 8B) and observe steady gains as the model becomes larger (SFT: $26.5\% \to 27.5\% \to 28.2\%$; DPO: $29.3\% \to 30.1\% \to 31.4\%$). Overall, both larger training data and larger model lead to better performance.

\section{Conclusion}
We presented a reasoning-augmented framework for next occupation prediction, leveraging oracle reasons as high-quality supervision to fine-tune LLMs. Our results show that incorporating explicit reasoning significantly improves prediction accuracy over zero-shot LLMs and traditional recommenders, and that jointly training reasoning and prediction yields the best performance. The study further confirms that better reasons, i.e., more factual, coherent, and utility-aligned, leads to better occupation predictions. Overall, our findings highlight the value of integrating reasoning into career trajectory modeling and suggest promising directions for building transparent, effective, and user-aligned occupation recommendation systems.

Although our study demonstrates the promise of reasoning-augmented LLMs for next-occupation prediction, several limitations remain.
First, our dataset is skewed towards US-based college-graduated career trajectories, limiting the generalization to other labor markets or demographic groups.
Second, our evaluation exclusively focuses on single-step next-occupation prediction. Multi-step or long-horizon career planning remains unexplored.
Third, the generation and evaluation of oracle reasons rely on an LLM-as-a-Judge paradigm. While effective, this approach means the quality of our supervised reasoning data is inherently dependent on the Judge LLM's own alignment and biases, as opposed to human expert validation.
Lastly, the use of fine-tuned LLMs introduces significant higher computational overhead and inference latency, compared to conventional sequential recommenders like SASRec. This approach, therefore, requires balancing the benefit of high interpretability against the drawback of increased computational cost.
Future work may address these limitations by extending to more diverse career data, exploring human-validated or richer forms of reasoning supervision, and developing efficient model compression techniques for fast inference.

\section*{Acknowledgment}
This research/project is supported by the Ministry of Education, Singapore, under its MOE Academic Research Fund Tier 2 programme (Proposal ID: T2EP20223-0047). Any opinions, findings and conclusions or recommendations expressed in this material are those of the author(s) and do not reflect the views of the MOE, Singapore.

\bibliographystyle{splncs04}
\bibliography{custom}

\clearpage

\end{document}